\documentclass[lettersize,journal]{IEEEtran}

\usepackage{amsmath,amsfonts}
\usepackage{algorithmic}
\usepackage{array}
\usepackage[caption=false,font=normalsize,labelfont=sf,textfont=sf]{subfig}
\usepackage{textcomp}
\usepackage{stfloats}
\usepackage{url}
\usepackage{verbatim}
\usepackage{graphicx}
\usepackage{cite}
\usepackage{tabularray} 
\usepackage{xltabular}
\usepackage{booktabs}
\usepackage[colorinlistoftodos, textsize=footnotesize, textwidth=35mm]{todonotes}
\usepackage[nolist]{acronym} 
\usepackage{hyperref} 
\usepackage{harveyballs}
\usepackage{multirow}
\usepackage{rotating}
\usepackage{multirow}
\usepackage{makecell}
\usepackage{duckuments}
\usepackage{float}
\usepackage{ulem}

\usepackage{epsfig}
\usepackage{subcaption}
\usepackage{calc}
\usepackage{amstext}
\usepackage{multicol}
\usepackage{pslatex}
\usepackage{algorithm2e}
\usepackage{siunitx}
\usepackage{colortbl}
\usepackage{array}
\usepackage[nameinlink, capitalise]{cleveref}
\usepackage{pgfplots}
\usepackage{svg}
\usepackage{adjustbox}
\usepackage{soul}
\usepackage{layouts}
\usepackage{gensymb}

\definecolor{whitesmoke}{rgb}{0.96, 0.96, 0.96}

\def\BibTeX{{\rm B\kern-.05em{\sc i\kern-.025em b}\kern-.08em
    T\kern-.1667em\lower.7ex\hbox{E}\kern-.125emX}}
\usepackage{balance}

\begin{document}

\title{vEDGAR - Can CARLA Do HiL?}
\author{Nils Gehrke¹, David Brecht¹, Dominik Kulmer¹, Dheer Patel¹, Frank Diermeyer¹
    \thanks{1  Munich Institute of Robotics and Machine Intelligence, Institute of Automotive Technology, Technical University of Munich, Corresponding author: Nils Gehrke (e-mail: nils.gehrke@tum.de)}
    }
\markboth{Journal of \LaTeX\ Class Files,~Vol.~XX, No.~X, November~2025}%
{How to Use the IEEEtran \LaTeX \ Templates}

\IEEEpubid{0000--0000/00\$00.00~\copyright~2021 IEEE}

\maketitle

\begin{abstract}
Simulation offers advantages throughout the development process of automated driving functions, both in research and product development.
Common open-source simulators like CARLA are extensively used in training, evaluation, and software-in-the-loop testing of new automated driving algorithms.
However, the CARLA simulator lacks an evaluation where research and automated driving vehicles are simulated with their entire sensor and actuation stack in real time.
The goal of this work is therefore to create a simulation framework for testing the automation software on its dedicated hardware and identifying its limits. 
Achieving this goal would greatly benefit the open-source development workflow of automated driving functions, designating CARLA as a consistent evaluation tool along the entire development process.
To achieve this goal, in a first step, requirements are derived, and a simulation architecture is specified and implemented.
Based on the formulated requirements, the proposed vEDGAR software is evaluated, resulting in a final conclusion on the applicability of CARLA for HiL testing of automated vehicles. 
The tool is available open source: Modified CARLA fork: \cite{TUMCARLARepo}, vEDGAR Framework: \cite{vEDGARRepo} \newline
\end{abstract}



\section{Introduction}

Simulation is a key element in the development of automated driving \cite{karle2024, Hu2024, Li2024, Son2019}. 
It is used to evaluate, optimize, and test software on the unit, interface, system, and hardware levels.
The necessity of simulation for automated driving systems can further be backed up by the amount of commercial simulation software available. 
Especially during the testing stage of the development process, simulation of both sensor data and vehicle dynamics is essential. 
While Software in the Loop (SiL) testing is already addressed with several simulation tools \cite{Geller+2024Carlos, Kaljavesi+2024CarlaAutowareBridge, Dosovitskiy+2017Carla, TierIV2024AWSIM}, an open-source tool for reliable Hardware in the Loop (HiL) testing is not yet available to the authors' knowledge.
Specifically, the CARLA simulator is currently deployed for SiL tests according to  \cite{Geller+2024Carlos, Kaljavesi+2024CarlaAutowareBridge}. Using CARLA as a foundation of the HiL simulation would achieve a consistent testing procedure.

\begin{figure}
    \centering
    \includegraphics[width=1.0\linewidth]{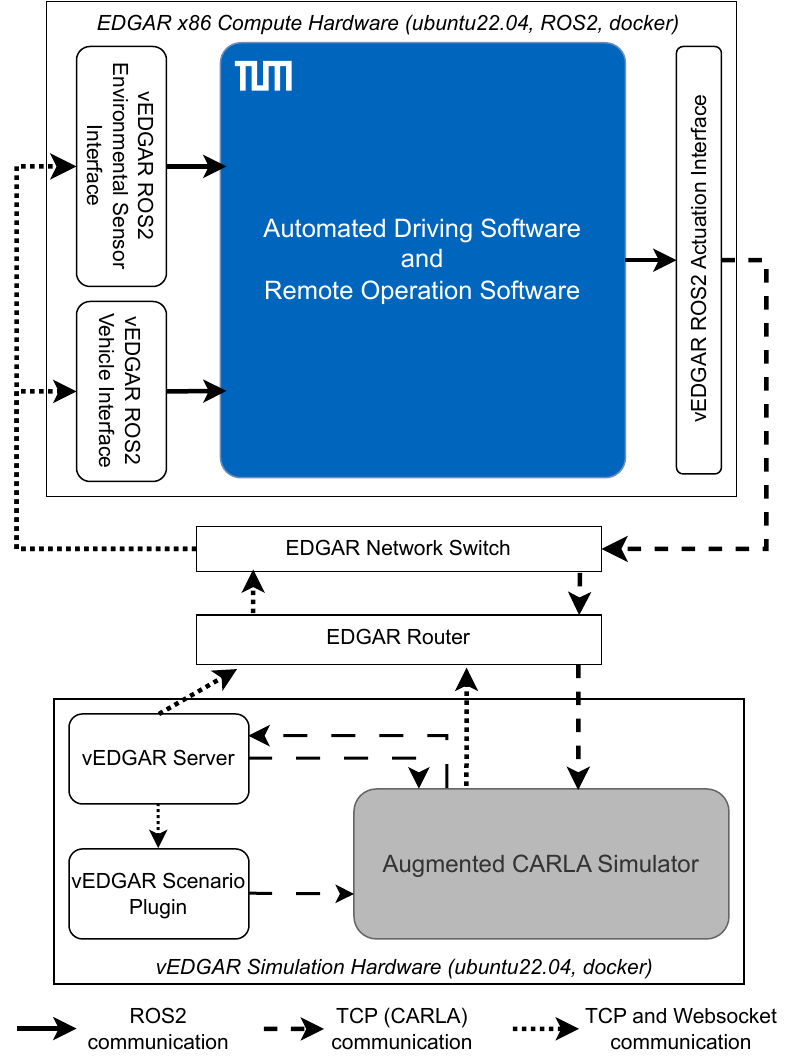}
    \caption{The HiL setup investigated in this work consists of the x86 compute hardware and network setup from the EDGAR research vehicle, together with the vEDGAR tool that wraps CARLA for the HiL application}
    \label{fig:HiLvEDGAR}
\end{figure}

To investigate the development of a HiL simulation tool using CARLA, this work aligns with the structured development process outlined in \cite{vmodell}. Essential steps, such as stakeholder analysis, requirement elicitation, implementation, and validation, are conducted. The target product of this development process shall be a modified CARLA version as well as a tooling software, further referred to as vEDGAR, that together form the simulation of the HiL in figure \ref{fig:HiLvEDGAR}.

\subsection{Definition of a Hardware-in-the-Loop}
\IEEEpubidadjcol
The HiL testing is a subcategory of the overall testing pipeline. Together with Model in the Loop (MiL) and Software in the Loop (SiL), the HiL represents an advanced stage of the testing process. The goal of a HiL simulation is to test a specific software within specific operational constraints on its designated hardware, typically ECUs. The HiL uses simulation to feed data into the potential input of the component under test. It closes the loop with the virtual environment by processing the output of the component. In the case of an automated driving function, the input can be described as sensor and CAN data from the vehicle, while the output is described as control commands to be executed by the vehicle \cite{Abel2006, Rosique2019}. 

\subsection{Detailed Problem Formulation for the EDGAR HiL}
Based on the HiL hardware layout provided in \cite{karle2024}, the proof of concept for this paper consists of constructing a component HiL for the automation software running on the x86 compute hardware in EDGAR. The ultimate goal is to construct a testing environment that allows for evaluating the automated driving software within the runtime constraints of the target hardware. This goal ultimately leads to a HiL abstraction specified at the ROS2 interface of the EDGAR automated driving software. This abstraction layer is defined in \ref{fig:edgarros2}. 

\begin{figure}
    \centering
    \includegraphics[width=0.9\linewidth]{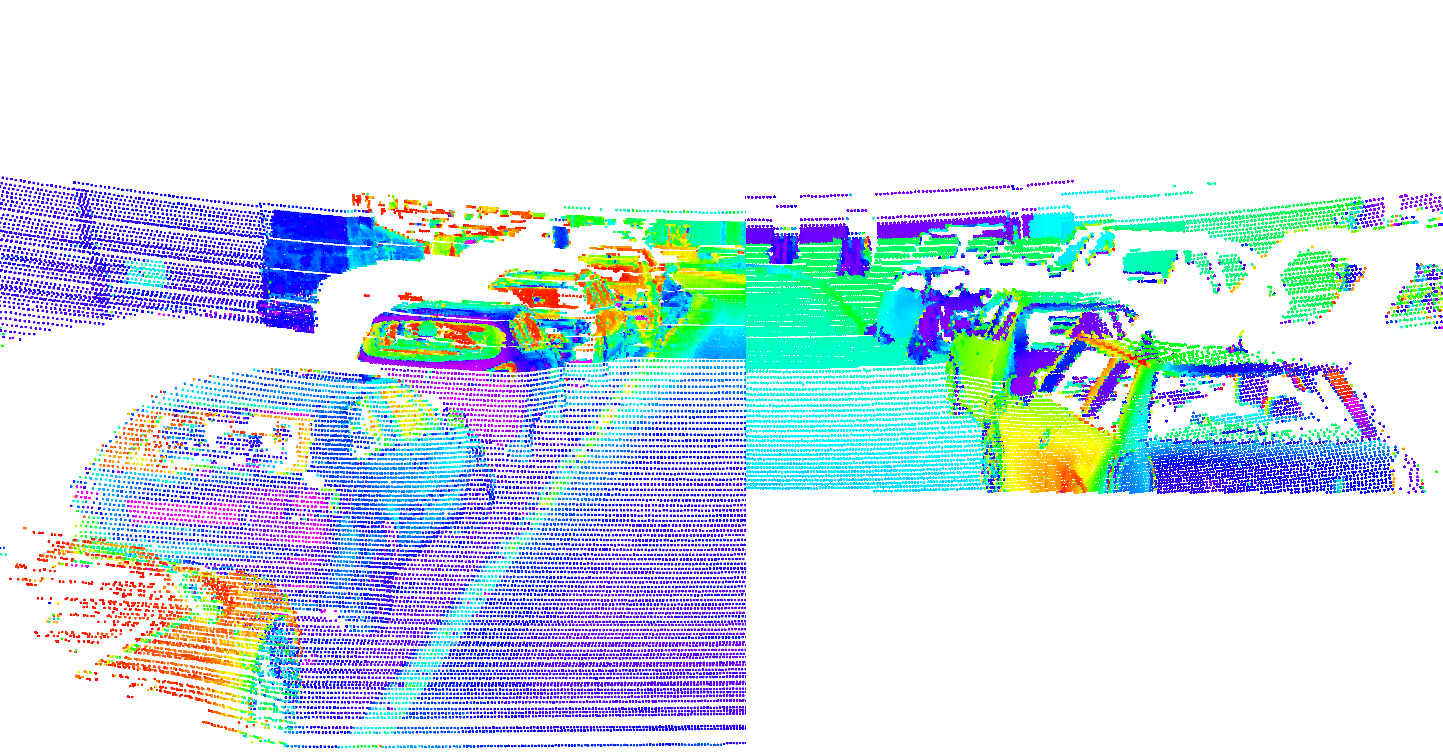}
    \caption{comparison of real LiDAR from EDGAR (left) and the simulated LiDAR from vEDGAR at the inner city of munich for the Wiesn-shuttle demo.}
    \label{fig:realworldapplication}
\end{figure}

The abstraction layer presents an optimal balance between the required development effort for research purposes and the benefits during testing. In the case of ECU testing, the hardware interfaces would be connected to the simulation software. especially with the custom EDGAR CAN interface and the different, individual sensors at EDGAR, this would be feasible only with large development resources. Ultimately, rather than providing the data input streams for the sensor and vehicle drivers within the HiL simulation, the ROS2 interface is mimicked, replacing the drivers with vEDGAR services, as indicated in Figure \ref{fig:HiLvEDGAR}. This approach allows for testing the automation software on the target hardware, with the limitation of neglecting driver-specific influence on the system performance. 
The application of the HiL software vEDGAR for real world application is indicated in figure \ref{fig:realworldapplication}, where the a copy of the real world track was created in the simulation to evaluate the system performance.

The simulation of individual sensors and actuation components occurs on dedicated simulation hardware, as described in \cite{karle2024}. The resulting network setup is thus different from the EDGAR network, and is only to a limited degree subject to the HiL testing. The layout from the switch towards the EDGAR remains the same. However, the VLAN configuration from the EDGAR research vehicle for different sensors is not applied.

\subsection{Characterization of the Stakeholders \label{subsec:stakeholder}}
According to \cite{vmodell}, stakeholders are all relevant bodies or people that have an interest in the planned project. The following section briefly introduces the relevant bodies interested in the use of HiL for Automated Driving Evaluation in research.

The identified stakeholders for this work are the research community, individual automated driving researchers, and the open-source community. Although these stakeholders present certain overlaps, each stakeholder introduces individual user stories and requirements for the usage of a HiL in the context of Figure \ref{fig:HiLvEDGAR}.
The user stories from the research community can be summarized according to:
\begin{itemize}
    \item[US1] The research community wants to evaluate multiple automated driving algorithms within a reproducible scenario on the HiL to comply with guideline 8 in \cite{DFG_Researchconduct}.
    \item[US2] The research community shares the evaluation approach and tools with other members of the community, e.g. via open source, to comply with guideline 10 in \cite{DFG_Researchconduct}.
    \item[US3] The research community wants to document the evaluation of automated driving algorithms at the HiL according to comply with guideline 12 in \cite{DFG_Researchconduct}.
\end{itemize}
In addition, the automated driving researcher proposes additional research stories:
\begin{itemize}
    \item[US4] The researcher wants to test a developed algorithm with little overhead on the HiL to have a benefit towards evaluating the algorithm directly on the vehicle.
    \item[US5] The researcher wants to use the HiL as an indicator that the developed algorithm will behave similarly on the EDGAR vehicle to reduce the debugging time on the vehicle.
    \item[US6] The researcher wants to configure the ROS2 interface of the vEDGAR to test new approaches that are currently not deployable on EDGAR (e.g., due to missing sensors).
    \item[US7] The researcher wants to reproduce scenes or scenarios observed in the reality to increase the robustness of the algorithm.
    \item[US8] The researcher wants to trigger critical scenarios in the simulation to evaluate the automated driving software safety without endangering vehicles or potentially damaging the EDGAR vehicle.
    \item[US9] The researchers want to consistently test their automated driving software stack in reproducible situations to create a baseline for future research. This also includes ground truth data from the submodules to compare intermediate steps in the automation software.
\end{itemize}
The open source community, on the other hand, inquires additional usage of the HiL setup tool:
\begin{itemize}
    \item[US10] the open source community wants to extend the existing HiL setup with additional tools for configuration or data output to fit it to new application areas.
    \item[US11] the open source community wants to repurpose the vEDGAR tool to use it as a HiL for their specific research vehicle.
\end{itemize}

As specialists, they would like to extend the existing software to meet their specific needs. 

\subsection{Requirement Formulation for the Usage of CARLA in a Research HiL} 
Based on the stakeholder specification in \ref{subsec:stakeholder}, in accordance with \cite{IEEE830-1998}, requirements are derived. These requirements can be used to assess the suitability of existing tools and, ultimately, to validate the proposed tool for its intended purpose. According to \cite{IEEE830-1998}, the requirements should cover the functions that the software should include, interface characterization, performance requirements, and other relevant attributes. Furthermore, the requirements should be verifiable and ranked by importance \cite{IEEE830-1998}. Requirements are derived from the previously presented user stories. An overview is provided in table \ref{tab:requirements}.

\begin{table}[!h]
	\caption{Main system requirements}
	\label{tab:requirements}
	\centering
	\begin{tblr}{stretch=1.2, colspec={Q[c,1.5cm]|Q[l,6.1cm]},rowspec={Q[m]Q[m]|Q[m]|Q[m]|Q[m]|Q[m]|Q[m]|Q[m]},row{1} = {bg=whitesmoke}}
		\hline[1pt]
		\textbf{Requirement}  & \textbf{Description} \\
		\hline[1pt]
		RQ-1 & \textbf{EDGAR ROS2 actuation interface:} The vehicle interface is provided via ROS2 [US1, US4, US7]\\
		RQ-2 & \textbf{Soft real time sensor interface:} The sensor information of EDGAR is made available in similar quantity and quality by the HiL [US1, US4, US7]\\
		RQ-3 & \textbf{Realistic closed loop behavior:} The vEDGAR shall provide reliable, similar closed loop behavior compared to the real vehicle [US5, US7]\\
		RQ-4 & \textbf{Resproducable vEDGAR configuration:} vEDGAR shall provide means to configure the ROS2 interfaces for research via a common configuration interface [US3, US7] \\
        RQ-5 & \textbf{Ground truth data:} The vEDGAR shall enable prototyping of AV modules and functions by providing ground truth data of intermediate steps in the sense-plan-act structure [US3, US4, US5, US7] \\
		RQ-6 & \textbf{Scenario configuration:} vEDGAR shall provide a easy-to-use scenario configuration in CARLA, including critical scenarios [US6, US7] \\
		RQ-7 & \textbf{Extendability:} vEDGAR shall be modular to enable future extension with limited implementation efforts [US2, US9]\\ \hline[1pt]
	\end{tblr}
\end{table}
 
Requirements RQ-1, RQ-2, and RQ-3 create a fundamental part of HiL testing and evaluation. They result consequently from the HiL context, according to Figure \ref{fig:HiLvEDGAR}. Requirements RQ-4, RQ-5, and RQ-6 enhance the usability of the tool, making it more versatile. With well-defined interfaces for manipulating the sensor interface or scenarios, as well as ground truth data for comparison, the applicability for research is increased. RQ-7 focuses on the open source community and enables extendability and future work with the vEDGAR tool for research and testing.

\subsection{Contribution}
In this work, the application of the simulation tool CARLA for Hardware in the Loop (HiL) testing and evaluation in automated driving research is investigated. 
This includes, in the first step, defining the requirements for the HiL simulation tool in a research context. 
In a second step, the current CARLA simulation tool is extended to meet the requirements. 
Lastly, the fulfillment of the requirements is evaluated. This evaluation is concluded by a proof-of-concept evaluation with varying compute loads, as well as an assessment of the suitability of the provided HiL setup for automated driving testing. 
With this work, the open-source tool vEDGAR \cite{vEDGARRepo, TUMCARLARepo} for automated driving software HiL testing is made available to the research community.
\section{State of the Art}

\subsection{HiL in Autonomous Driving Research}
The HiL is a versatile tool to test hardware and software together in a simulated environment. 
\cite{Abel2006} and \cite{Rosique2019} define the HiL as a setup that allows testing the functionality of the final hardware using a simulation of external inputs and outputs on dedicated simulation hardware. 
Commonly, the HiL is processed by a Model-in-the-Loop (MiL) and Software-in-the-Loop (SiL) during the development process \cite{Rosique2019}.
The usage of a HiL in literature is demonstrated  \cite{Deng2008, Gelbal2017, Rosique2019, Sievers2018}.
Furthermore, the Sense-Plan-Act architecture is employed to structure the application of the HiL across various sources.

\cite{Deng2008} and \cite{Gelbal2017} evaluate the control behavior of their functions on a HiL.
\cite{Deng2008} applies multiple vehicle models for the testing of the autonomous driving system. Notable in this approach is the incorporation of steering actuation hardware into the HiL. Based on a designed scenario, the behavior of an Adaptive Cruise Control is analyzed, and the velocity profile of the leading and following vehicles is compared.
\cite{Gelbal2017} tests a cooperative cruise control and a lane-keeping controller in a HiL. Simulation is performed via a Simulink model and data provided by a CAN bus. The HiL is used to analyze the behavior of the control for a specific input scenario, both for lane keeping and cruise control.
Both examples also simulate high-level sensor input via Simulink \cite{Gelbal2017}\cite{Deng2008}.

For testing perception algorithms, a dedicated sensor model is required for each simulated sensor. 
\cite{Rosique2019} summarizes the available simulation tools for different sensor types, including versatile simulation tools such as CARLA \cite{Dosovitskiy+2017Carla} and dedicated solutions for LiDAR, Radar, and Sonic sensors.
Overall, game engines are well established for sensor simulation, benefiting from already available functions and high performance in visual rendering. The authors in \cite{Rosique2019} conclude the importance of high-fidelity sensor simulation for the testing of automated driving perception functions. 
They further emphasize the benefits of a HiL for rapid prototyping. \cite{Deng2008} demonstrates high-level fusion of sensor modalities in a HiL.

\cite{King2019} highlights the importance of driving scenarios in HiL testing in their framework concept. 
The authors emphasize the need for test catalogues to evaluate automated driving functions. 
Such testing involves an ontology for describing the test, which validates the completeness of the test cases.
Additionally, the activation criterion of specific tests needs to be defined.
Finally, the results shall be measured according to specific test conditions \cite{King2019}. 
The literature confirms the validity of the requirements RQ-1 to RQ-6. A suitable open-source solution for an EDGAR HiL is not presented.

\subsection{Extensions of the CARLA Simulator}
Due to its broad usage, CARLA and its available open source extensions are compared against the defined requirements and user stories.
CARLA \cite{Dosovitskiy+2017Carla} is a widely used open-source simulation tool based on Unreal Engine, offering multiple assets and a comprehensive Python API.
Due to its well-documented API and extensibility, CARLA serves as the foundation for numerous tools dedicated to advancing research in automated driving. While featuring extensive sensor and vehicle dynamics simulation, CARLA 0.9.15 does not natively support ROS2. This has changed with the latest release of CARLA 0.9.16, which, however, is not investigated in this work. Furthermore, the Radar and LiDAR models lack a degree of fidelity, which is also due to their runtime not being compatible with the RQ-2. RQ-5 and RQ-6 are only feasible via extensive usage of the Python API. RQ-7 can be achieved via the Python API or the C++ client library.

The CARLA ROS Bridge \cite{CarlaSimulator2022CarlaRosBridge} comprises of a C++ library that parses interfaces from the CARLA API to ROS2 Messages or Services packages, enabling communication between CARLA and ROS2. While addressing the missing ROS2 compatibility, additional ROS2 nodes are necessary to comply with RQ-1 for the EDGAR vehicle.
Building on the CARLA ROS Bridge, the CARLA Autoware Bridge \cite{Kaljavesi+2024CarlaAutowareBridge} enables the integration of CARLA and the open-source automated driving software Autoware, further extending the utility of the simulation framework in automated driving research. Via this tool,  RQ-1 is still not fully resolved; however, the gap has narrowed further.
Yet, since it builds on the CARLA ROS Bridge, it comes with the same shortcomings as the overall CARLA simulator.

The CARLA scenario runner is a tool for the CARLA simulator that enables the definition of scenarios within CARLA. Such scenarios can either be defined via OpenScenario or via custom Python files \cite{ScenarioRunner}. As an extension of the CARLA simulator, it features suitable means to define and play repetitive scenarios in CARLA. 
Additionally, the open-source CARLOS framework \cite{Geller+2024Carlos} leverages the CARLA platform using the CARLA ROS Bridge and the CARLA scenario runner to provide a simulation framework designed to facilitate the integration and testing of ROS2 applications within the domain of connected intelligent transportation systems.

\subsection{Research Vehicle EDGAR}
To establish a clear reference for comparison and to parameterize vEDGAR in terms of vehicle-specific attributes such as sensor positions, sensor types, and vehicle dynamics, this section introduces the EDGAR research vehicle.
While the presented tool is vehicle-agnostic and can adapt to various automated driving platforms and sensor configurations, EDGAR serves as a representative baseline.

\begin{figure}[!t]
    \centering
    \noindent\includegraphics[width=1.0\linewidth]{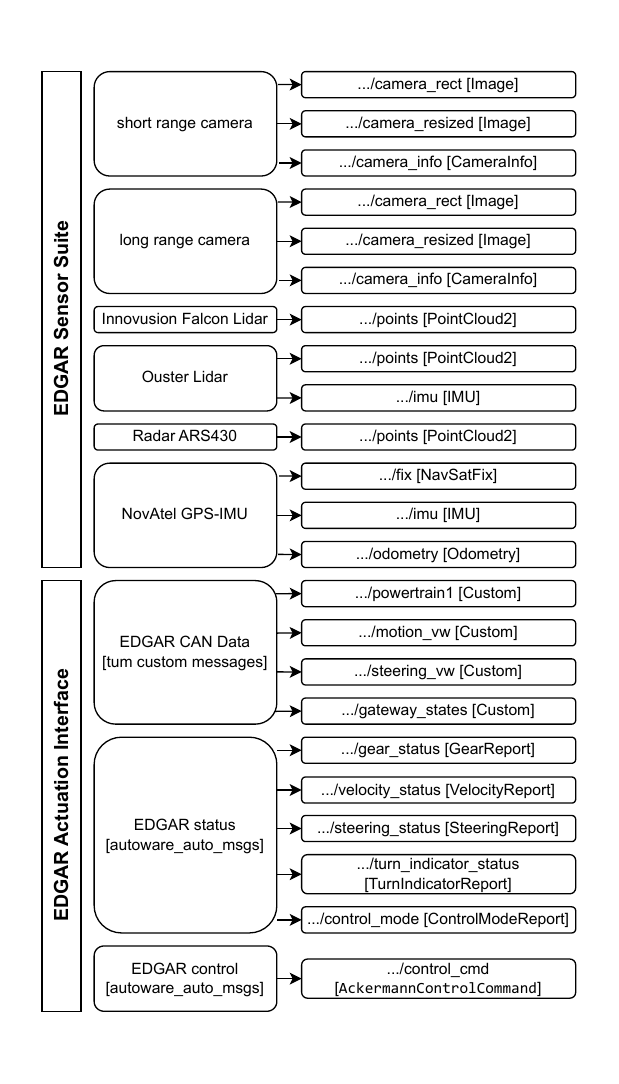}
    \caption{Schematic representation of the interfaces of the research vehicle EDGAR, [] indicates the ROS2 message class}
    \label{fig:edgarros2}
\end{figure}

EDGAR is based on a Volkswagen T7 Multivan equipped with an array of sensors, actuator interfaces, and computational hardware. 
A schematic representation of its interfaces is provided in \cref{fig:edgarros2}.
The sensor suite includes, among others, short- and long-range cameras, LiDAR, Radar, GNSS, and IMU sensors, all of which support ROS2 interfaces.
For a comprehensive inventory of the installed sensors, refer to \cite{karle2024}.
The actuator interface enables control via input of desired steering angles and acceleration values.
The basis of the automation system used for benchmarking in this paper is implemented using Autoware. The overall EDGAR automated driving system, including the described sensor stack and automated driving software, was successfully deployed in public traffic through multiple real-world tests \cite{WiesnShuttle, IAA2025}, further enhancing the suitability of the platform as a proof of concept for a CARLA HiL.

\subsection{Research Gap}
The provided tools in the literature already partially address the posed requirements in \ref{tab:requirements}. 
RQ-1 needs detailed modelling of the EDGAR interface.
Requirement RQ-2 is not addressed by any of the proposed tools but remains essential for the HiL application.  Missing in all approaches is also a validation of RQ-3 by comparing it to real-world vehicle behavior. RQ-4, RQ-5, and RQ-6 need to be addressed further, with RQ-4 and RQ-6 benefiting from a Graphical Interaction Interface for configuration. Overall, the applicability of CARLA within a HiL for automated driving systems remains uncertain in the state of the art, and will be addressed in this work based on the EDGAR research vehicle.
\section{Making CARLA HiL-Ready}
\begin{figure}[!t]
    \centering
    \noindent\includegraphics[width=1.0\linewidth]{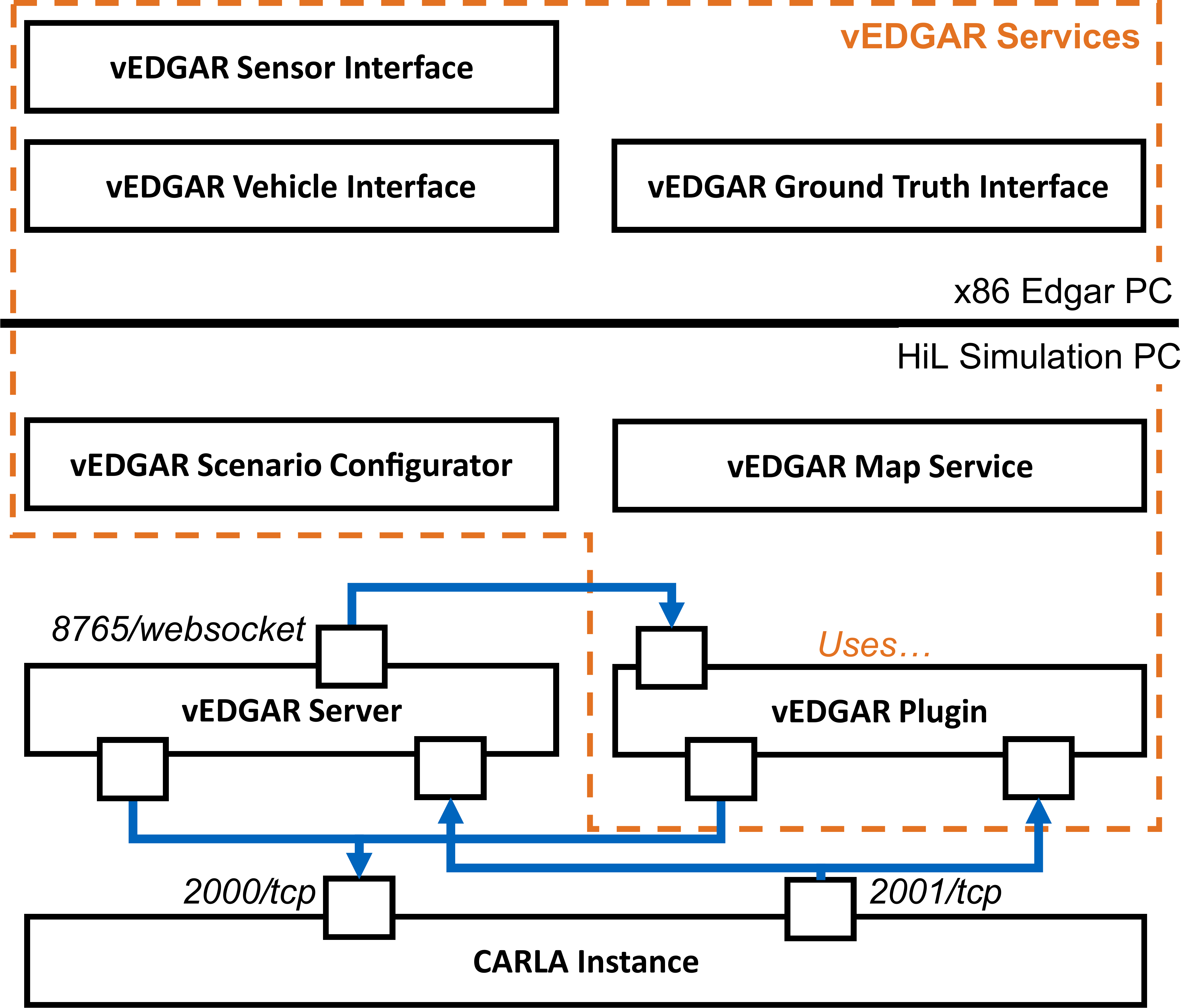}
    \caption{vEDGAR architecture applied for the HiL. Displayed are the various services that utilize a vEDGAR Plugin instance, the vEDGAR Server, and the CARLA Server, as well as how these services are distributed across the hardware. Blue arrows describe a data transmission via the network.}
    \label{fig:vedgar_architecture}
\end{figure}
Based on the state of the art, using the CARLA simulation for automated driving HiL testing requires two major adaptations. Firstly, the CARLA simulator must be accelerated to fulfill RQ-2 in combination with an increase in sensor fidelity. Furthermore, the additional tool, vEDGAR, based on the CARLA Python API, is introduced to address Requirements RQ-1 through RQ-7. 

\subsection{vEDGAR Architecture}
Figure \ref{fig:vedgar_architecture} presents the selected service-oriented architecture for the vEDGAR HiL software. Using a vEDGAR server and individual vEDGAR Plugins, RQ-7 is addressed. 
The vEDGAR server handles the general simulation configuration for the HiL. It is responsible for spawning the EDGAR research vehicle actor in CARLA, enforces a real-time step in synchronous mode, and provides the simulation details, including the CARLA simulation IP and the EDGAR actor ID, to the plugins via a websocket.  

The vEDGAR Plugin serves as the counterpart to implement individual services. From the requirements, 5 major services for the vEDGAR are extracted:
\begin{itemize}
    \item \texttt{vehicle\_interface\_service}
    \item \texttt{sensor\_interface\_service}
    \item \texttt{scenario\_configurator\_service}
    \item \texttt{groundtruth\_publisher\_service}
    \item \texttt{map\_service} .
\end{itemize}
All services inherit the vEDGAR Plugin class and run independently from each other. Only services related to ROS2 interfaces run on the x86 PC, replacing the EDGAR sensor drivers. This split of services is possible due to the intermediate vEDGAR server.

This architecture in figure \ref{fig:vedgar_architecture} aligns with RQ-7. New features do not require code changes at other services, the vEDGAR server, or the plugin, and can be developed and deployed independently of one another. 

\subsection{Soft Real-Time Sensor Kit}
Real-time computation, as described in \cite{Shin1994}, consists of three major characteristics. 
First, the software should be able to return results for an input before a specified deadline. 
Further, the software needs to be reliable and, as a third characteristic, also robust with regard to external operation conditions. 
The used CARLA version is based on Unreal Engine 4. 
This game engine does not fulfill these hard real-time requirements, according to \cite{Shin1994}.

To still use the powerful functionalities of the CARLA simulator, further, a soft real-time constraint is discussed: 
The HiL simulation is considered soft real-time capable if the same time passes in real time and simulation during one simulation step. This is ensured by an internal timer. Furthermore, the same frame rates as the equivalent real sensors shall be achieved.
While the vehicle physics in CARLA fulfill this weakened real-time condition, the required full EDGAR sensor stack remains a challenge in the current CARLA simulation. The following steps are applied to address RQ-2, RQ-4, and RQ-6. 
\subsubsection{Porting to a GPU-LiDAR}
\begin{figure*}
    \centering
    \includegraphics[width=1.0\linewidth]{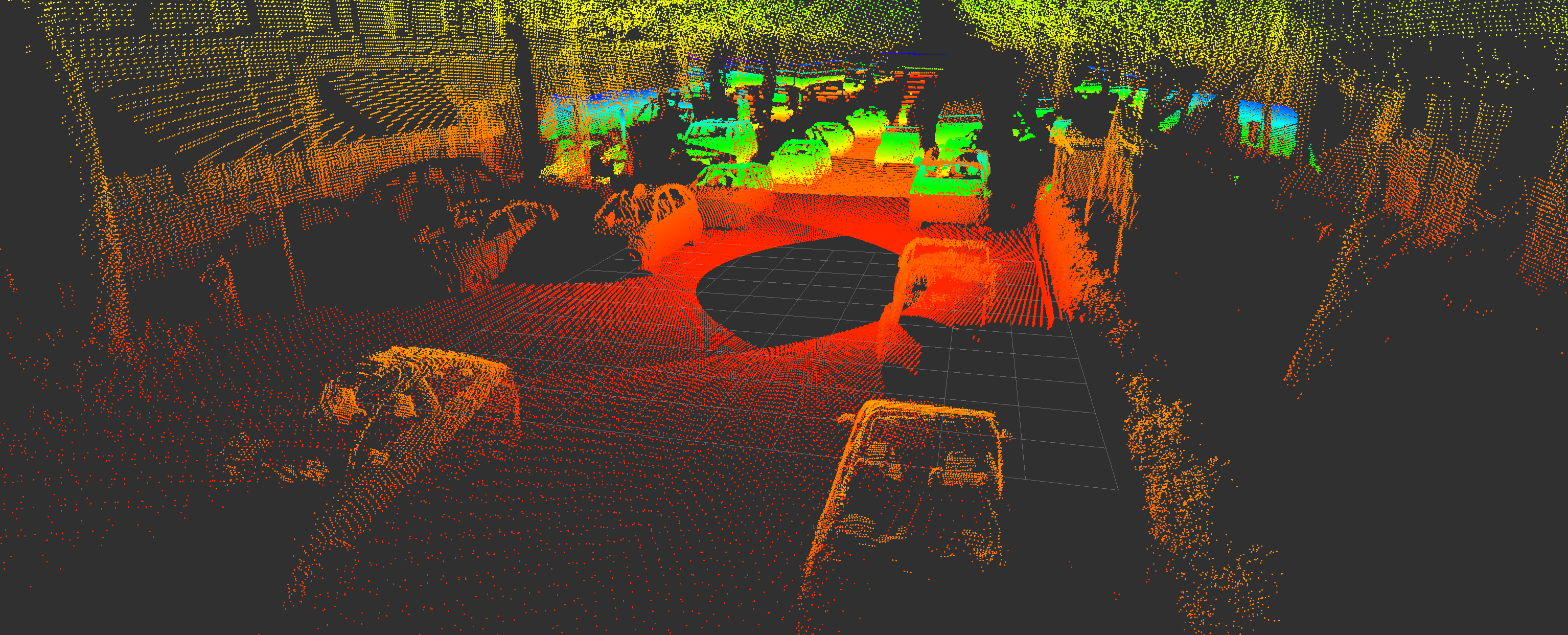}
    \caption{The final LiDAR simulation achieved runs with 10Hz in a soft real-time condition using the GPU acceleration. It consists of four individual LiDAR sensors, two mechanical Ouster OS1 with 360 degrees coverage and two solid state Innovusion Falcon with a FoV of 120 degrees. The concatenated point cloud of the four LiDARs is visible in the image. Each sensor is simulated individually and merged in the automation softwarestack.}
    \label{fig:lidarscan}
\end{figure*}

An analysis of the EDGAR sensor kit according to \cite{EdgarDigitalTwin} concludes that the CARLA LiDAR sensor remains the only sensor failing the soft real-time constraint.
This was already observed in \cite{Kaljavesi+2024CarlaAutowareBridge}.
The reduced performance originates from a CPU-based ray tracing of the sensor. This results in high computation times for high-resolution LiDARs.
A soft real-time LiDAR can be achieved by porting the sensor simulation to the GPU.
Based on the Unreal Scene Capture Components, LiDAR points are sampled from an online-calculated depth texture. 
The rays and their corresponding pixels are either precomputed from a configured LiDAR resolution or are loaded directly from a scan pattern. 
Based on the projection matrix, the point is sampled from the corresponding pixel z-buffers. This approach is especially suited for solid-state LiDARs with a limited field of view. Values above 170 degrees are approximated by multiple capture components.
The simulation runtime increases significantly for the EDGAR LiDAR Sensors, achieving up to 16Hz and contributing to RQ-2. The final results for the EDGAR LiDAR Scan are depicted in figure \ref{fig:lidarscan}.

\subsubsection{ROS2 Publisher Factory and Fail Modes}
The specific sensor service uses a sensor factory to construct individual ROS2 SensorNodes based on a YAML preconfiguration. The individual sensor configurations are provided in two stages, a first yaml defines the SensorTypes containing the CARLA blueprint parameters for the sensors, a second file defines the transformation, topic and frame id of the sensor. Both files together address RQ-4. The sensor interface service then constructs the individual CARLA sensors and ROS2 wrappers via sensor-specific factories that are collected and defined in a FactorySelector class.
Benefit of this approach is the straight forward extendability for further custom sensors that might be based on new or existing carla sensors, e.g a stereo camera. Additionally, multiple CARLA sensors, such as a GNSS and IMU, can be combined into a single research vehicle sensor in this approach. Further, fail modalities like sensor failures can be defined at the SensorNode level and scaled on the entire sensor kit.
Additionally, the transform tree \texttt{tf\_static} is automatically generated and provided by the sensor service based on the provided YAML files. A GUI allows toggling specific sensors at runtime, e.g., simulating their sudden failure. This contributes to RQ-6.
A detailed list of a subset of preconfigured sensors is given in \cite{vEDGARRepo} and summarized in table \ref{tab:sensor_kit}. The parameters are based on the CARLA Python API.

\begin{table}[!h]
    \centering
    \begin{tblr}{stretch=1.2, colspec={Q[c, 2.1cm]|Q[l,6.1cm]},rowspec={Q[m]Q[m]|Q[m]|Q[m]|Q[m]|Q[m]|Q[m]|Q[m]|Q[m]|Q[m]|Q[m]},row{1} = {bg=whitesmoke}}
		\hline[1pt]
		\textbf{Sensor}  & \textbf{Parameters} \\
		\hline[1pt]
        Basler Camera Front & \texttt{image\_size\_x=960}, \texttt{image\_size\_y=600}, \texttt{sensor\_tick=0.05}, \texttt{fov=84.9}, \texttt{iso=100}, \texttt{gamma=2.2}, \texttt{focal\_distance=6000} \\     
        Basler Camera Rear & \texttt{image\_size\_x=960}, \texttt{image\_size\_y=600}, \texttt{sensor\_tick=0.05}, \texttt{fov=99.5}, \texttt{iso=100}, \texttt{gamma=2.2}, \texttt{focal\_distance=6000} \\     
        Basler Camera Long Range & \texttt{image\_size\_x=800}, \texttt{image\_size\_y=600}, \texttt{sensor\_tick=0.1}, \texttt{fov=38.6}, \texttt{iso=100}, \texttt{gamma=2.2}, \texttt{focal\_distance=16000} \\     
        Innovusion Falcon LiDAR & \texttt{horizontal\_fov=150}, \texttt{vertical\_fov=40}, \texttt{horizontal\_resolution=0.1}, \texttt{vertical\_channels=152}, \texttt{sensor\_tick=0.1}, \texttt{range=250}, \texttt{x\_standard\_deviation=0.001}, \texttt{y\_standard\_deviation=0.001}, \texttt{z\_standard\_deviation=0.001}, \texttt{scan\_pattern\_path="/path/to/pattern.csv"} \\     
        Ouster OS1 LiDAR & \texttt{horizontal\_fov=360}, \texttt{vertical\_fov=45}, \texttt{horizontal\_resolution=0.351}, \texttt{vertical\_channels=128}, \texttt{sensor\_tick=0.1},\texttt{range=100} \\
        
    \end{tblr}
    \caption{Sensor kit configuration in vEDGAR according to \cite{karle2024} and calibration measurements. The configurations represent the EDGAR sensor interface.}
    \label{tab:sensor_kit}
\end{table}

\subsection{Realistic Actuation Interface for Closed Loop}
CARLA enables closed-loop simulation via control interfaces for vehicle actors. In detail, two input modalities are provided: an Ackermann control and a vehicle control with throttle, brake, handbrake, and steering inputs. These interfaces are used and refined for RQ-3. Furthermore, the vehicle's internal sensor output, as outlined in the comprehensive overview in \ref{fig:edgarros2}, addresses RQ-1. Both contribute to the vehicle interface service.
The output of the vehicle interface service includes different reports from the \texttt{autoware\_msgs} as well as custom CAN messages from the EDGAR vehicle. While the first part is generic and can be applied to any vehicle type, the second part is specifically designed for the EDGAR vehicle. 
Data sources from CARLA are the last applied control command in CARLA, as well as vehicle states for the indicators. The gear is simplified through the two gears, Drive and Reverse, with initialization in the Drive position. This corresponds with the EDGAR vehicle.
Vehicle data is published at 50~\si{\Hz}, despite the lower refresh rate of the CARLA simulator, to adequately represent the EDGAR vehicle interface.

\begin{figure*}
    \centering
    \includegraphics[width=1.0\textwidth]{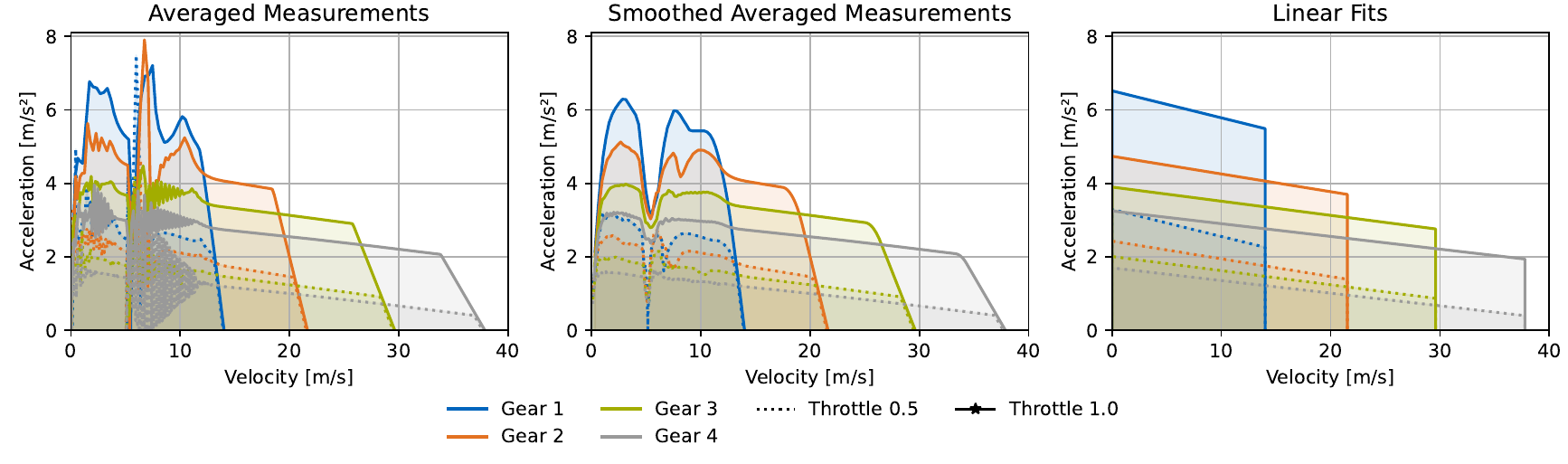}
    \caption{Measurements for static throttle inputs averaged over 1000 runs. The noisy acceleration near 5m/s is clearly visible. The linear fits are used for the control model later and perform well for static control inputs.}
    \label{fig:throttle_acceleration_mapping}
\end{figure*}

Control command input is received via a ROS 2 subscriber using the Autoware Ackermann control commands. The control commands are filtered and restricted to the allowed maximum steering angle and maximum acceleration from \cite{EdgarDigitalTwin}. They are then, in a first approach, forwarded as CARLA Ackermann control commands.
Initial observations indicate unstable control behavior at velocities exceeding 30~\si{km/h}. This behavior can be reproduced outside the vEDGAR framework at different velocities.
To address these instabilities, a second approach utilizes the CARLA vehicle interface, which includes definitions for throttle, brake, and steering angle. A mapping between a current velocity, a target acceleration, and the respective throttle for stationary throttle situations, as displayed in figure \ref{fig:throttle_acceleration_mapping}, is suggested. 
Together with a PI feedback controller, the target acceleration is transformed into a desired throttle output. Despite exhibiting stable behavior for stationary acceleration inputs during testing, it requires careful tuning of the PI in combination with the automated driving system.

The final approach for the closed loop consists of a straightforward integration of the input acceleration over time. This target velocity is then applied within the physics calculation. This results in a precise application of the desired acceleration. When remaining within an acceleration range of $a \in [-3.0, 3.5]$, the modeled assumption of an error-free execution of the target acceleration is considered suitable for the EDGAR HiL testing. All introduced control modes are selectable within vEDGAR. The used interface, however, should be the subject of future research, further contributing to RQ-3.

\subsection{Sense-Plan Ground Truth Data for Prototyping}
RQ-5 directly requests rapid prototyping of AV modules within the vEDGAR HiL framework.
The requirement directly relates to ROS2 ground truth publishers for major interfaces within the Sense-Plan-Act structure based on the Autoware architecture \cite{autoware}. A respective vEDGAR service thus provides all surrounding objects as detected objects, their past positions as tracked objects, and the predicted objects to evaluate or replace the Sense Module. 

This data is extracted from CARLA. The detected objects are directly accessible via the list of actors in the scene, along with their bounding boxes and types. 
Tracking data is obtained by collecting past position, orientation, and velocity of all actors in the scenes.
The predicted trajectory requires additional logic. In the case of the CARLA actor being controlled by a CARLA traffic manager, the planned path can be extracted from the traffic manager. Otherwise, waypoints from the road network are used.
A similar approach can be applied to construct a potential trajectory for the planning module.

\subsection{Scenario Configuration for the Edgar HiL}
Crucial to HiL testing is the ability to create and test scenarios that align with RQ-3 and RQ-4. Together with the RQ-5 to utilize the vEDGAR toolset for prototyping and the RQ-7 for extendability, vEDGAR introduces additional means for scenario configuration that extend the existing CARLA Scenario Runner \cite{ScenarioRunner}. The CARLA Scenario Runner can, in addition to other features, play OpenScenario files within CARLA.
Additionally, the CARLA Python API offers the following configuration options: placing static vehicles on the map, controlling vehicles on the map via the traffic manager, spawning pedestrians on the map, allowing them to walk in a predefined direction, and placing static props. Further, traffic lights and weather can be controlled.

The scenario service aims to provide a simple and reproducible interface for CARLA functions, eliminating the need to use the Python API.
The six-layer model splits a scenario into six different categories \cite{Weber2019}. In the following, layers 3 to 5 are addressed by the vEDGAR service via a comprehensive GUI. 
Layer 3 is achieved by placing static vehicles or props along a visualization of the OpenDrive road network. The static vehicles are oriented according to the nearest road point.
This enables the straightforward configuration of parking vehicles, second-row parkers, and lane-blocking vehicles.
Layer 5 is achieved by providing the CARLA weather parameters in a GUI.

Layer 4 contains multiple features in the vEDGAR scenario service. In the first step, CARLA-controlled vehicles can be placed along the OpenDrive network using the mouse. The vehicle's further actions are then determined via a random seed.
To generate high traffic areas in city environments, vehicle sources and sinks can be placed by the user. Sources continuously spawn new vehicles within a user-defined time delay, while sinks destroy all actors passing through the indicated position. The EDGAR vehicle is not affected by the sinks. This enables rapid configuration of dense traffic at specific intersections or roads on the map.
Additionally, the user can configure pedestrian movement via the lanelet. By providing a pedestrian path, a crowd size, and a respawn delay, pedestrians are continuously spawned in the simulation and complete the indicated path. This mode enables both creating pedestrian interactions at intersections and addressing critical scenarios where pedestrians suddenly cross roads outside dedicated pedestrian crossings, as per RQ-6. 
The scenario configurations can be saved and loaded in accordance with RQ-4. Notably, vEDGAR remains compatible with the CARLA scenario runner, which can serve as a tool for more complex scenarios \cite{ScenarioRunner}.

\subsection{HMI Interfaces for the User}
\begin{figure*}
    \centering
    \subfloat[Sensor Configuration]{%
        \includegraphics[height=5cm]{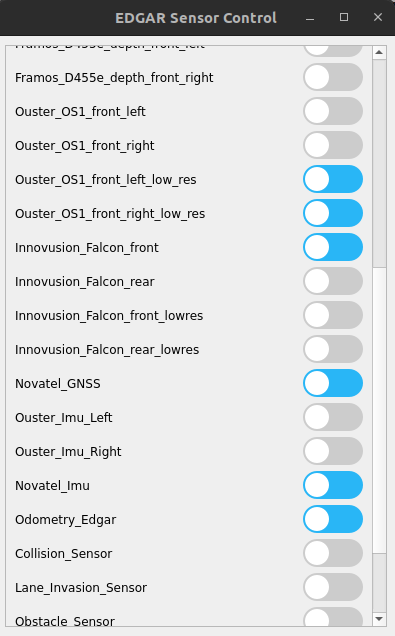}
    }\hfill
    \subfloat[Scenario Configuration]{%
        \includegraphics[height=5cm]{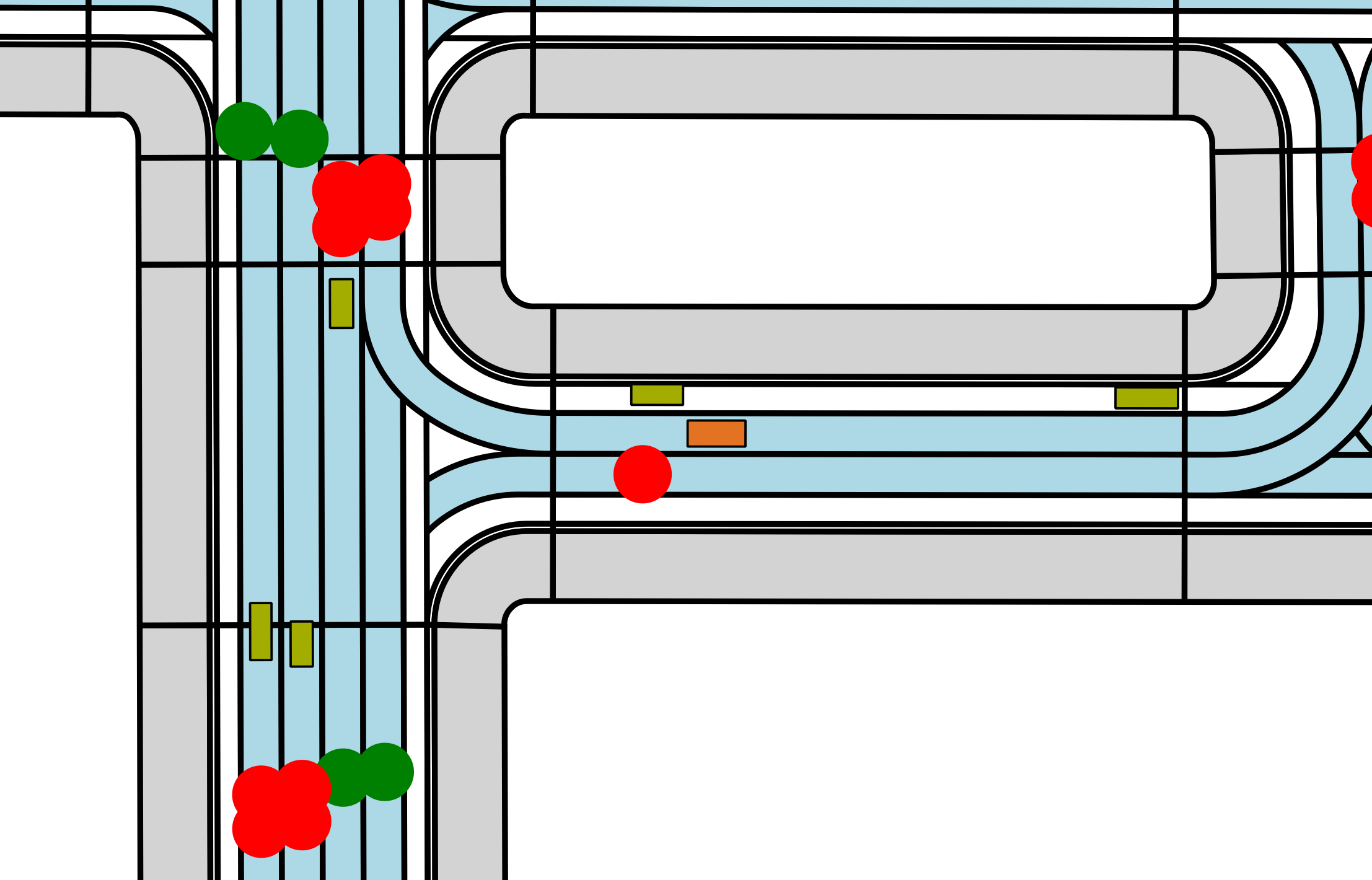}
    }\hfill
    \subfloat[Control Debug Output]{%
        \includegraphics[height=5cm]{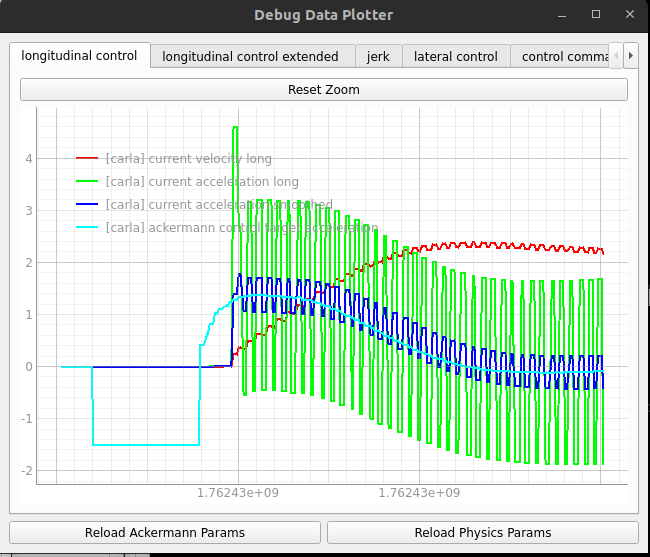}
    }
    \caption{This graphic depicts an excerpt of the three interfaces. a) shows the sensor configuration interface, b) the scenario configuration with exemplary spawn and deletion points, and c) the vehicle debug window.}
    \label{fig:gui}
\end{figure*}

The vEDGAR consists of multiple graphical user interfaces to facilitate usage. These interfaces enable configuring the sensor kit and sensor failure, debugging the EDGAR control input through input and control execution monitoring, and configuring scenes via an interactive OpenDrive map.
The respective GUI interfaces are presented in Figure \ref{fig:gui}.

\subsection{Enabling Distributed Simulation on multiple PCs}
Scalability of the simulation environment on different hardware is crucial for HiL testing with large sensor setups. 
By fulfilling the soft real-time condition, each sensor is individually able to provide its data bandwidth via the CARLA simulator. 
The entirety of the sensor setup, however, cannot always be simulated via a single CARLA simulation instance with the available hardware.
One CARLA simulation instance is currently restricted to one hardware GPU. 
To use multiple CARLA simulation instances, CARLA offers a multi-GPU setup. This setup runs and synchronizes one CARLA simulation instance per GPU on one physical compute device. 

The drawback of this approach is that it requires the simulation to run in synchronous mode. In synchronous mode, when running in a soft real-time condition, the same computation time is enforced for all sensors. Given a camera running at 20Hz and a LiDAR running at 10Hz, both the LiDAR simulation and the camera simulation would have 50 milliseconds to complete their simulation step. This can result in undesired behavior, namely, if the LiDAR is not finished within 50 milliseconds. Consistent observations revealed buffering of the simulation calls, and consequently, a delayed LiDAR data stream. The stream, however, is provided with the latest timestamp and thus becomes very difficult to filter during a simulation run.
Furthermore, all synchronized CARLA instances in the multi-GPU setup require the same IP address, which limits the maximum GPU count with the current available hardware.

To address this problem, a synchronization mechanism is implemented for asynchronous simulation servers. Based on an Euler scheme, the reference actor position and velocity for server A are extrapolated to the current timestep on server B and then applied together with the latest control commands of the actor. 
Investigations of higher-order time schemes for extrapolation did not show noticeable improvements in the position accuracy of the synchronized CARLA actors in server B.
Server B now allows the spawning of dedicated sensors in the simulation, allowing better scalability for large sensor setups. Furthermore, the asynchronous mode addresses the previously described undesired behavior of simulation call buffering by allowing the sensor simulation to complete its current calculation.
The current implementation limits the number of synchronized actors to 20. However, this limitation could be increased in future work by utilizing the CARLA C++ interface.

\subsection{Additional Tools for Enhanced Usage}
Another crucial point of deploying an automated vehicle in reality is the respective high-definition map. In the case of the Autoware framework, this map consists of a fused point cloud scan, a lanelet2 map, and metadata files. 
In reality, a time-consuming process, including SLAM \cite{KulmerSLAM} and creating a lanelet2 map, is involved in creating such maps. In CARLA, the majority of the information is already encoded in the virtual environment. 
To reduce testing efforts on different maps, the point cloud scan can be created automatically from a CARLA map using the mapping service. While not directly contributing to the requirements, it addresses US11, US10, and US2. The corresponding lanelet2 map can be generated using existing tools, such as \cite{mai2021}.
\section{Results}
The fulfillment of requirements RQ1 to RQ7, as outlined in Table \ref{tab:requirements}, is evaluated through a multi-stage process.
In the experiments, two automated driving systems are used: a fully automated driving software \cite{karle2024} and a remote operation software \cite{Kerbl2025}, which is evaluated on the EDGAR hardware.
The evaluated success criteria are summarized in \ref{tab:evaluation_strategy}.

\begin{table}[!h]
    \centering
    \begin{tblr}{stretch=1.2, colspec={Q[c,1.5cm]|Q[l,6.1cm]},rowspec={Q[m]Q[m]|Q[m]|Q[m]|Q[m]|Q[m]|Q[m]|Q[m]|Q[m]|Q[m]|Q[m]},row{1} = {bg=whitesmoke}}
		\hline[1pt]
		\textbf{Requirement}  & \textbf{Condition for Successfull Validation} \\
		\hline[1pt]
        RQ-1 & The required ROS2 vehicle interface for the automated driving software under test is available. \\
        RQ-2 & The required ROS2 sensor data for the automated driving software under test are provided with the same bandwidth and refresh rate as in EDGAR.   \\
        RQ-3 & The vEDGAR allows mission completion evaluation via a closed loop.\\
        RQ-4 & The configuration of the EDGAR sensor setup according to \cite{karle2024} is successfully performed (see \ref{tab:sensor_kit}). \\
        RQ-5 & The difference between the ground truth perception data and the automated driving software stack is determined. \\
        RQ-6 & The configuration of a schematic critical scenario involving vulnerable road users is performed in vEDGAR only.\\
        RQ-7 & - \\
        System-1 & The fully automated driving system solves a reference track for different dynamic environment complexities. Changes in system behavior can be related to hardware demands. \\     
        
    \end{tblr}
    \caption{Success conditions for the HiL requirements}
    \label{tab:evaluation_strategy}
\end{table}

The evaluation of the success criteria is performed on the CARLA Town10 and a custom CARLA map at the Forschungscampus Garching.
The vEDGAR tool for point cloud map generation is used to create the respective Autoware maps for the automated driving software.
The hardware used for the simulation comprises three systems. Two desktop PC with an Intel 14900K Processor, 64GB of RAM, and an RTX 4090 are used for evaluations with a reduced sensor setup or as a secondary server for the HiL. The HiL simulation PC, equipped with an AMD Threadripper 7965, 128GB of RAM, and three RTX 4090 GPUs, is used to analyze the driving system's performance under different environmental complexities.

\subsection{Evaluation of the vEDGAR ROS2 interface}
To evaluate the provided ROS2 interface, the first step is to store all available ROS2 topics at EDGAR. These topics are then filtered to include those required by at least one of the two automated driving software stacks. In a third step, they are cross-checked with the available ROS2 topics from the vEDGAR interface.
The results are presented in Table \ref{tab:ROS2_Interface_eval}, with similar topic names grouped into single rows to enhance readability.

\begin{table}[!h]
    \centering
    \begin{tblr}{stretch=1.2, colspec={Q[l, 4.1cm]|Q[c,0.7cm]|Q[c,0.7cm]|Q[c,1.1cm]},rowspec={Q[m]Q[m]|Q[m]|Q[m]|Q[m]|Q[m]|Q[m]|Q[m]|Q[m]|Q[m]|Q[m]|Q[m]|Q[m]|Q[m]|Q[m]|Q[m]|Q[m]|Q[m]|Q[m]|Q[m]|Q[m]|Q[m]|Q[m]|Q[m]},row{1} = {bg=whitesmoke}}
		\hline[1pt]
		\textbf{ROS2 Topic} & \textbf{AD} & \textbf{RD} & \textbf{vEDGAR} \\
		\hline[1pt]
        /clock & X &   & \harveyBallFull \\
        /control/command/control\_cmd & X & X & \harveyBallFull \\
        /control/command/gear\_cmd & X & X & \harveyBallNone \\
        /control/command/...\_cmd & X & X & \harveyBallNone \\
        /edgar/can/.. & X & X & \harveyBallHalf \\
        /edgar/sensor/gnss/.../center/imu & X & X & \harveyBallNone \\
        /edgar/sensor/gnss/.../center/odom &   & X & \harveyBallFull \\        
        /edgar/sensor/gnss/... & X &   & \harveyBallNone \\
        /edgar/sensor/lidar/.../.../points & X & X & \harveyBallFull \\
        /edgar/sensor/.../.../.../camera\_info & X & X & \harveyBallFull \\
        /edgar/sensor/.../.../.../image\_rect & X &   & \harveyBallFull \\
        /edgar/sensor/.../.../.../image\_resized &   & X & \harveyBallFull \\
        /tf & X & X & \harveyBallNone \\
        /tf\_static & X & X & \harveyBallFull \\
        /vehicle/..\_engage & X &   & \harveyBallNone \\
        /vehicle/doors/status & X &   & \harveyBallNone \\
        /vehicle/sensor/fix & X & X & \harveyBallFull \\
        /vehicle/sensor/imu1 & X &   & \harveyBallFull \\
        /autoware\_orientation & X &   & \harveyBallFull \\
        /vehicle/status/steering\_status & X & X & \harveyBallFull \\
        /vehicle/status/gear\_status & X & X & \harveyBallFull \\
        /vehicle/status/velocity\_status & X & X & \harveyBallFull \\
        /vehicle/status/... & X &   & \harveyBallHalf \\
        \hline[1pt]
    \end{tblr}
    \caption{Available ROS2 Interface at vEDGAR. AD and RD abbreviate Automated Driving and Remote Driving, respectively, referring to the two driving systems being tested. The availability in vEDGAR is indicated by the Harvey balls in the categories: not available~\harveyBallNone, partially available~\harveyBallHalf, and fully available~\harveyBallFull.}
    \label{tab:ROS2_Interface_eval}
\end{table}

\subsection{Evaluation of the Soft Real-Time Sensor Kit}
According to \ref{tab:evaluation_strategy}, the soft real-time condition is evaluated by comparing the provided data rate and bandwidth of the EDGAR sensors with the vEDGAR sensors.
To achieve this, a ROS2 node subscribes to one of the sensor topics at a time, saving the topic size, the time since the last update, and the resolution in cases of images or point clouds. During the test, EDGAR remains in a stop position. Initially, each sensor is analyzed individually. The resulting comparison is provided in table \ref{tab:sensor_real-time_individually}. The data distinguishes between synchronous and asynchronous modes in CARLA. In the first mode, the vEDGAR framework triggers a new simulation step at a set time interval.
In an asynchronous simulation, CARLA selects the internal time step size dynamically based on the current simulation load. The time passed in the simulation can be considered equivalent to the time passed in the real world.

\begin{table}[!h]
    \centering
    \begin{tblr}{stretch=1.2, colspec={Q[c, 2.1cm]|Q[l,1.7cm]|Q[l,1.7cm]|Q[l,1.7cm]},rowspec={Q[m]Q[m]|Q[m]|Q[m]|Q[m]|Q[m]|Q[m]|Q[m]|Q[m]|Q[m]|Q[m]},row{1} = {bg=whitesmoke}}
		\hline[1pt]
		\textbf{Sensor}  & \textbf{Edgar} & \textbf{vEDGAR (sync)} & \textbf{vEDGAR (async)}  \\
		\hline[1pt]
        Basler Camera raw & 50ms, 2.3 MB & 110ms, 6.9MB & -\\    
        Basler Camera resized & 50ms, 1.4 MB & 52ms, 1.4 MB & 61ms, 1.4MB\\    
        Innovusion LiDAR &  100ms, 2.9MB & 160ms, 5.8MB & 126ms, 5.8MB\\     
        Ouster LiDAR & 100ms, 2.1MB & 130ms, 1.9MB & 113ms, 1.6MB\\     
        Ouster IMU & 10ms, 340 Byte & 100ms, 340 Byte & 100ms, 340 Byte \\     
        NovAtel IMU & 3ms, 332 Byte & - & 17ms, 324 Byte\\     
        NovAtel GNSS & 50ms, 141 Byte & 100ms, 141 Byte & 100ms, 141 Byte\\
        NovAtel Odometry & 10ms, 724 Byte & 50ms, 716 Byte & 50ms, 716 Byte \\ 
    \end{tblr}
    \caption{Comparison of the provided data between the EDGAR vehicle and the vEDGAR HiL software averaged over 4000 measurements.}
    \label{tab:sensor_real-time_individually}
\end{table}
The displayed results reflect both the internal CARLA computation time and the processing time via the vEDGAR Python services.

In a second run, the entire sensor setup is launched on the EDGAR and vEDGAR HiL Hardware. In this run, in addition the CPU and RAM load due to the sensor kit on the EDGAR Hardware is analyzed both on EDGAR and the HiL. The respective results are depicted in \ref{tab:sensor_real-time_kit}.

\begin{table}[!h]
    \centering
    \begin{tblr}{stretch=1.2, colspec={Q[l, 2.0cm]|Q[c,1.2cm]|Q[c,1.2cm]|Q[c,1.2cm]|Q[c,1.2cm]},rowspec={Q[m]Q[m]Q[m]|Q[m]|Q[m]|Q[m]|Q[m]|Q[m]|Q[m]|Q[m]|Q[m]Q[m]Q[m]Q[m]|Q[m]|Q[m]|Q[m]|Q[m]|Q[m]|Q[m]|Q[m]|Q[m]|Q[m]},row{1, 12} = {bg=whitesmoke}}
    \hline[1pt]
    \SetCell[c=5]{c} \textbf{EDGAR driver compute resources (averaged)} \\
    \hline[1pt]
    \textbf{name} & \textbf{cpu cores used [\%]} & \textbf{cpu time [s]} & \textbf{memory [\%]} & \textbf{memory [Mb]} \\
    \hline[1pt]
     camera           &          5.95 &         7390.39 &      0.49 &         627.34 \\                
     lidar innovusion &         13.66 &          604.47 &     0.12 &         152.11  \\
     lidar ouster     &         13.66 &          604.47 &      0.12 &         152.11  \\
     GNSS             &          9.33 &          314.99 & 0.04 &          51.60  \\
     actuation        &          9.81 &          356.13 & 0.08 &         106.53  \\
     bridge           &          2.22 &           82.55 & 0.05 &          60.51  \\
     can              &          4.76 &          197.10 & 0.04 &          53.15  \\
     state publisher  &          0.13 &            9.75 &  0.04 &          47.05  \\
     $\sum$           &      59.52 & 9559.85 & 0.98 & 1250.40\\
    \hline[2pt]
    \SetCell[c=5]{c} \textbf{vEDGAR ROS2 interface compute resources (averaged)} \\
    \hline[1pt]
    \textbf{name} & \textbf{cpu core used [\%]} & \textbf{cpu time [s]} & \textbf{memory [\%]} & \textbf{memory [Mb]} \\
    \hline[1pt]
     clock        &          0.39 &            7.56 &             0.04 &          54.03 \\
     sensors  &        251.81 & 3503.79	& 1.49	& 1920.37\\
     vehicle      &         20.97 &          291.19 &             0.09 &         112.55 \\
    $\sum$       &      273.17 &            3802.54 &               1.62 &       2086.95\\
    \end{tblr}
    \caption{A list of averaged compute resources required by the research vehicle drivers and HiL software for comparison. Docker stats were used to obtain the data.}
    \label{tab:sensor_real-time_kit}
\end{table}

\subsection{HiL Testing with vEDGAR}
In the final step, a proof-of-concept evaluation of the CARLA HiL Setup (vEDGAR) is used to determine potential degradations of the automated driving software due to hardware limitations. 
The HiL Setup for this experiment consists of the three previously introduced simulation PCs. These PCs run one CARLA instance per graphics card, which are included in the vEDGAR Framework as secondary servers. The Automated Driving Software, based on the Autoware software stack, is deployed on the vehicle's target hardware.

\begin{figure}
    \centering
    \includegraphics[width=0.5\textwidth]{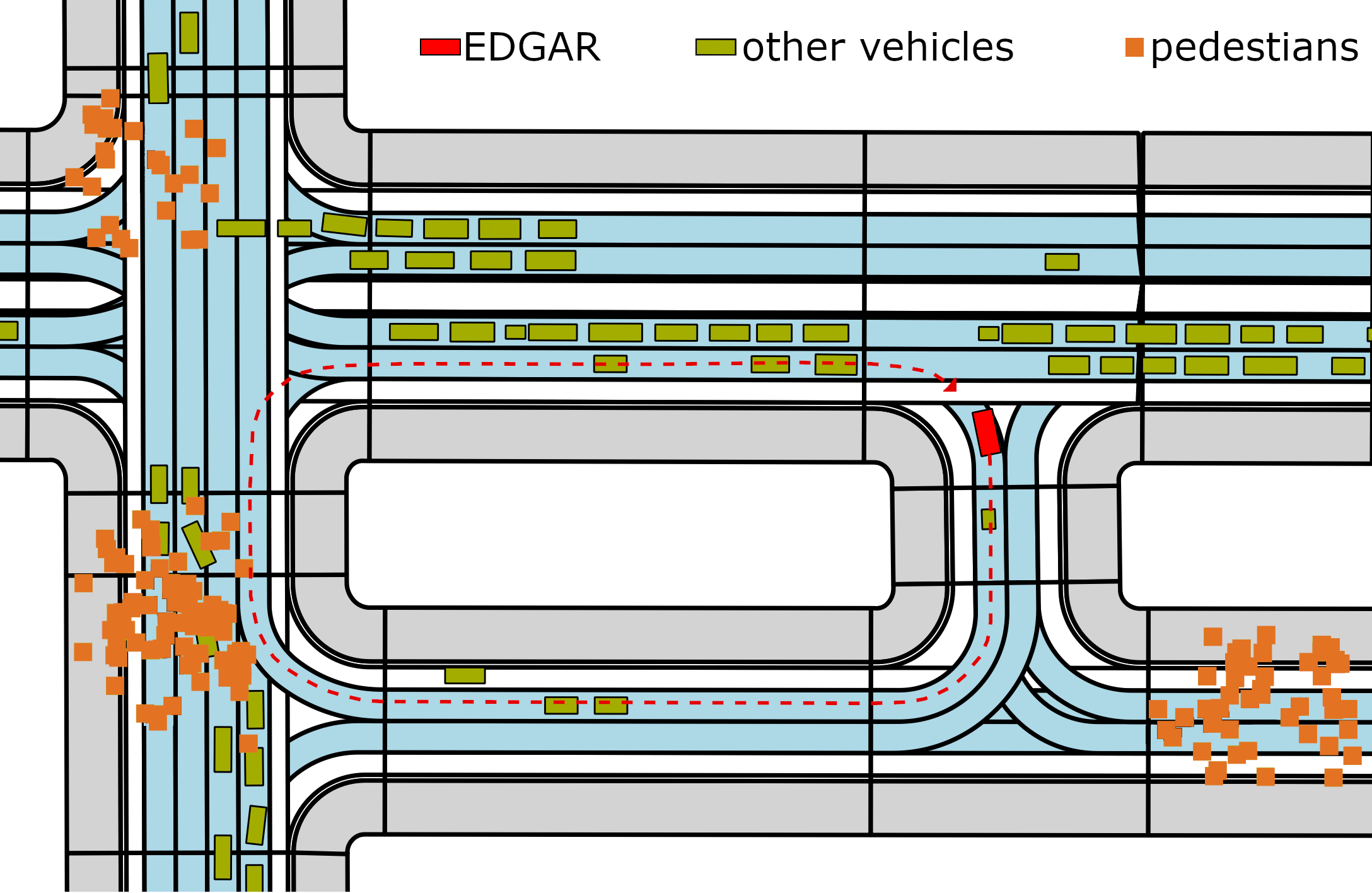}
    \caption{Configuration of the three scenes in the vEDGAR scenario configurator service. The complexity increases from scene one to three. In each scene, EDGAR follows the red line performing multiple loops. Depending on the scene, other road users or pedestrians are part of the environment, moving dynamically in accordance with the road network.}
    \label{fig:scenarios_HiL_PoC}
\end{figure}

\begin{figure*}
    \centering
    \includegraphics[width=1.0\textwidth]{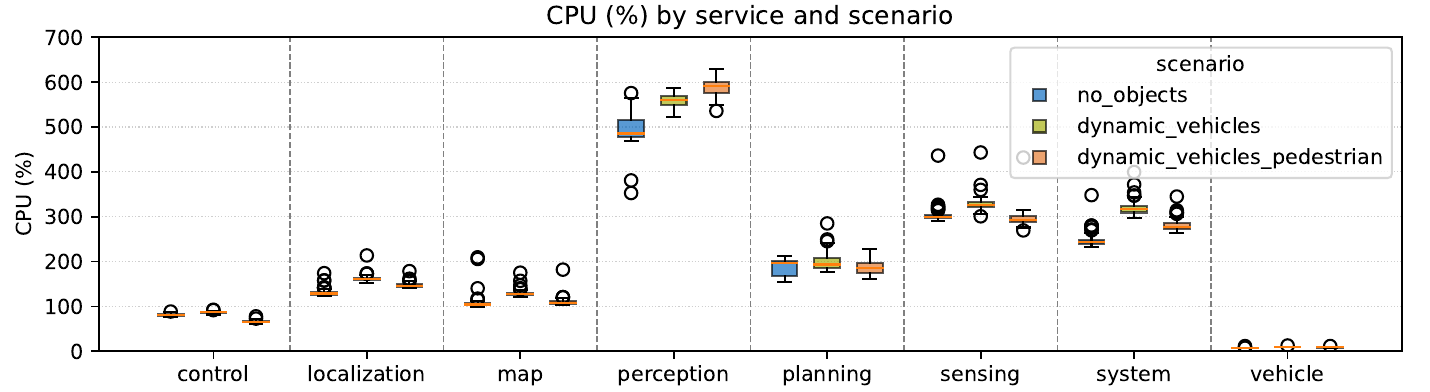}
    \caption{Used CPU cores per software module on the target hardware.}
    \label{fig:final_eval}
\end{figure*}

To test the impact of increasing compute demands inside the automation software on the driving performance, the three different scenes in figure \ref{fig:scenarios_HiL_PoC} are executed. 
The automation software is instructed to always follow the same lane course around a building block in Town10, consisting of right turns and straights. To model different computation demands, the dynamic environment of the vehicle is altered. In a first run, no other road users are present on the map. The second run places spawn points around the intersections, resulting in a crowded city environment with numerous dynamic vehicles. The third run additionally adds pedestrian crowds at specific areas around the target track. The pedestrians are crossing the road. They are placed such that they do not cross the target lane of the automated driving vehicle. For all scenarios, the compute demand of the automation software is measured. Additionally, the executed trajectory is recorded and compared. The evaluation of the different modules of the driver stack shows only minor differences in the compute demand for the perception and prediction, as displayed in figure \ref{fig:final_eval}. The target hardware, despite the complex situations, does not limit the performance. This is further confirmed via the driven trajectories, which show no significant difference.

\section{Discussion}
The results of the evaluation display a fulfillment of the defined requirements. All requirements can be considered fulfilled through the proof-of-concept evaluation with the automated driving software stack. The authors, however, conclude from the results that the resulting CARLA implementation together with vEDGAR is operating at its limit when used for a HiL. Notably, CARLA is not intended for this use case. Slight improvements are estimated when switching to CARLA 0.9.16. Due to their dedicated evaluation, RQ-2 and RQ-3 are discussed in detail.

Results from the evaluation of RQ-2 in table \ref{tab:sensor_real-time_individually} and \ref{tab:sensor_real-time_kit}, however, show discrepancies between the EDGAR and vEDGAR ROS2 data rate and bandwidth and the utilized compute resources. This can be partially explained by the usage of Python for the data conversion from CARLA to ROS2 and the buffering problem in synchronous mode. Furthermore, the refresh rate was evaluated using a ROS2 node, which may introduce additional uncertainties in the measured data rates by skipping individual ROS2 topics due to the QoS settings of the sensor messages. Evaluations of the CARLA internal LiDAR Sensor simulation, which presented itself as the most critical towards RQ-2, present an average computation time of around 50~ms per point cloud. The sensor models align with the current state of the art. However, they still do not fully represent the real-world sensor behavior. Nevertheless, crucial contributions towards RQ-2 are made and evaluated.

RQ-3 showed challenges within the physics simulation of the CARLA simulator during the implementation phase. Observing the vehicle motion from a macroscopic scale presents a stable physics behavior. Applying target velocities via the CARLA Ackermann control interface is observed to behave stably. The noisy acceleration output of the simulation, however, renders a stable acceleration-based control interface of the vehicle difficult. Model-based knowledge was applied to reduce the effect of the noisy acceleration on the vEDGAR internal acceleration control. The most stable and robust internal vehicle control mode turned out to be, however, the simple integration of target accelerations into a target velocity. This approach, however, leaves room for improvement and presents an idealized model of the EDGAR vehicle interface.

Further, only a limited set of scenes and test rides is used for evaluation. Due to the nature of the sensor implementation logic, runtime evaluations for RQ-2 are not considered to change greatly when considering more simulation runs, a doubling in data samples from 4000 to 8000 showed no noticable changes in average frame rate or data bandwidth. Especially, the RQ-6 could benefit from additional simulation runs. This could help to further clarify the limits of the proposed basic scenario configuration and create guidelines on when to switch to the CARLA scenario runner.

\section{Conclusion}

In this work, the application of CARLA for an automated driving HiL is investigated. The specific proof of concept evaluated is a component HiL of the EDGAR compute hardware with a reduced representation of the EDGAR network. To evaluate the HiL framework, three stakeholders were identified and seven requirements derived.

To meet the requirements, changes to the CARLA simulator, as well as the development of the vEDGAR tool, are necessary. For the vEDGAR tool, a service-based architecture is proposed. Major contributions of this work include a GPU-accelerated LiDAR, a modular simulation architecture, and a configurable ROS2 interface. Further, the specific design of the vehicle actuation interface is investigated, comparing three different approaches.

The final evaluation of the requirements showed a general applicability of CARLA. However, it was at the limits of its capabilities. The main drawbacks of the CARLA simulator, primarily its inability to support a hard real-time condition, could not be fully countered by accelerated sensor implementation and offloading to different compute PCs. Nevertheless, testing two  automated and remote driving software stacks using the vEDGAR tool is consistently achieved for multiple runs. Overall, the created tool can be alternatively used on a single compute machine with a reduced simulation-to-real-time ratio to test automated driving stacks outside the HiL context. 

Future work includes, but is not limited to, migrating the vEDGAR tool and the custom CARLA sensor to CARLA 0.9.16 with native ROS2 support, further improving the actuation interface model, and conducting a scenario-based evaluation using the vEDGAR framework, such as the CARLA scenario runner. The tool is available at \cite{vEDGARRepo} and \cite{TUMCARLARepo}.

\section*{Acknowledgments}
Nils Gehrke, as the first author, contributed the initial idea and did the majority of the work. David Brecht and Dominik Kulmer provided crucial input to the structure and reviewed the manuscript critically. Dheer Patel reviewed the manuscript and the open source tools. The authors would like to thank Walim Grira, who contributed to the initial implementation of synchronizing different CARLA servers as part of his bachelor's thesis. Frank Diermeyer critically revised the manuscript for its intellectual content and provided final approval for the published version. He agrees with all aspects of the work.
The research was partially funded by the Bavarian Research Foundation (BFS), the Federal Ministry of Research, Technology and Space (BMFTR) and through basic research funds from the Institute of Automotive Technology (FTM). 

In this work, the tool ChatGPT was used for code debugging. The tool Grammarly was used for proofreading the document. The final document and code were carefully reviewed by the authors for correctness.

\bibliographystyle{IEEEtran}
\bibliography{bibliography}

\vspace{11pt}

\begin{IEEEbiography}[{\includegraphics[width=1in,height=1.25in,clip,keepaspectratio]{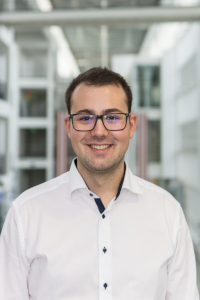}}]{Nils Gehrke} received the B.Sc. degree in mechanical engineering from ETH Zürich, and the M.Sc. degree in computational science and engineering from the Technical University of Munich, where he is currently pursuing the Ph.D. degree as a Scientific Researcher with the Institute of Automotive Engineering and his research focuses on safety of automated vehicles. He is currently leading the simulation team for the EDGAR research vehicle.
\end{IEEEbiography}
\begin{IEEEbiography}
[{\includegraphics[width=1in,height=1.25in,clip,keepaspectratio]{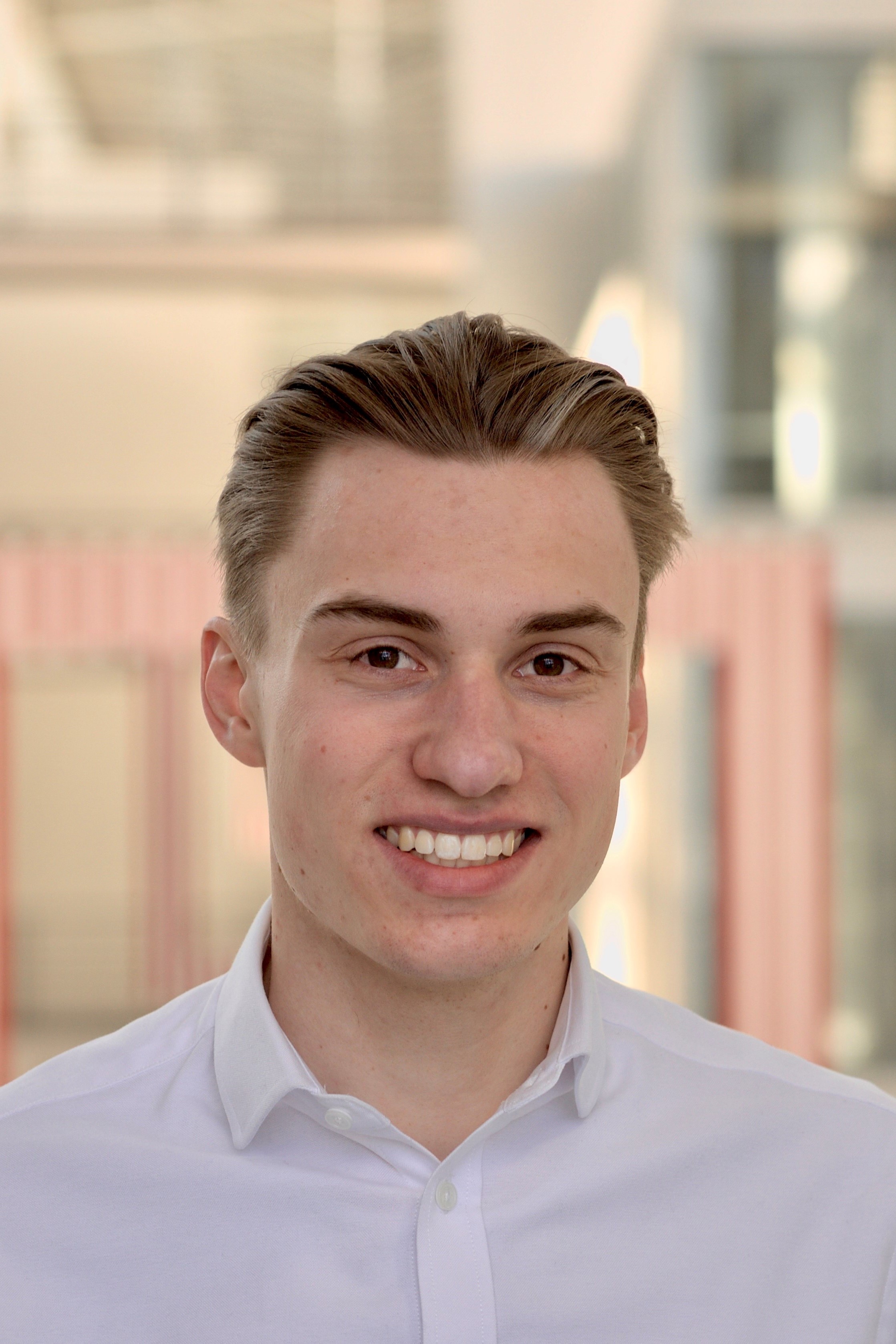}}]{David Brecht} received the B.Sc. degree in mechanical engineering and the first M.Sc. degree in automotive engineering from Technische Universität Braunschweig in 2020 and 2022, respectively, and the second M.Sc. degree in automotive engineering from Tongji University in 2023. He is currently pursuing the Ph.D. degree in mechanical engineering with the Institute of Automotive Technology, Technical University of Munich. His research interests include safety of teleoperated and automated vehicles.
\end{IEEEbiography}
\begin{IEEEbiography}
[{\includegraphics[width=1in,height=1.25in,clip,keepaspectratio]{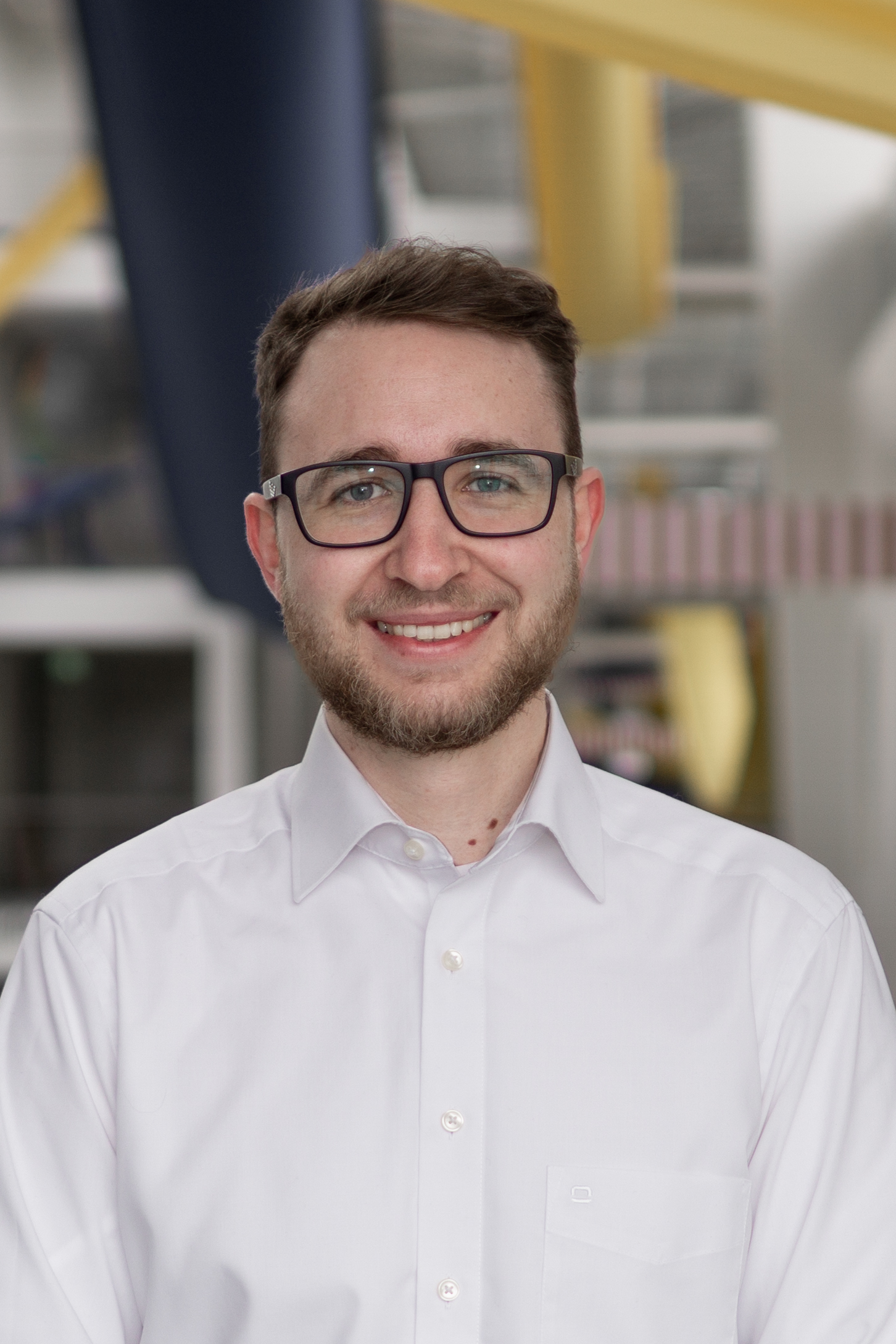}}]{Dominik Kulmer} received a B.Sc. degree in mechanical engineering and an M.Sc. degree in ~automotive engineering from the Technical University of Munich in 2020 and 2022, respectively. He is currently pursuing a Ph.D. degree in automotive engineering with the Institute of Automotive Technology, Technical University of Munich. His research focuses on georeferenced localization and mapping in GNSS-denied environments for high-speed autonomous racecars and for public road traffic. He is currently a co-lead of the EDGAR project.
\end{IEEEbiography}
\begin{IEEEbiography}
[{\includegraphics[width=1in,height=1.25in,clip,keepaspectratio]{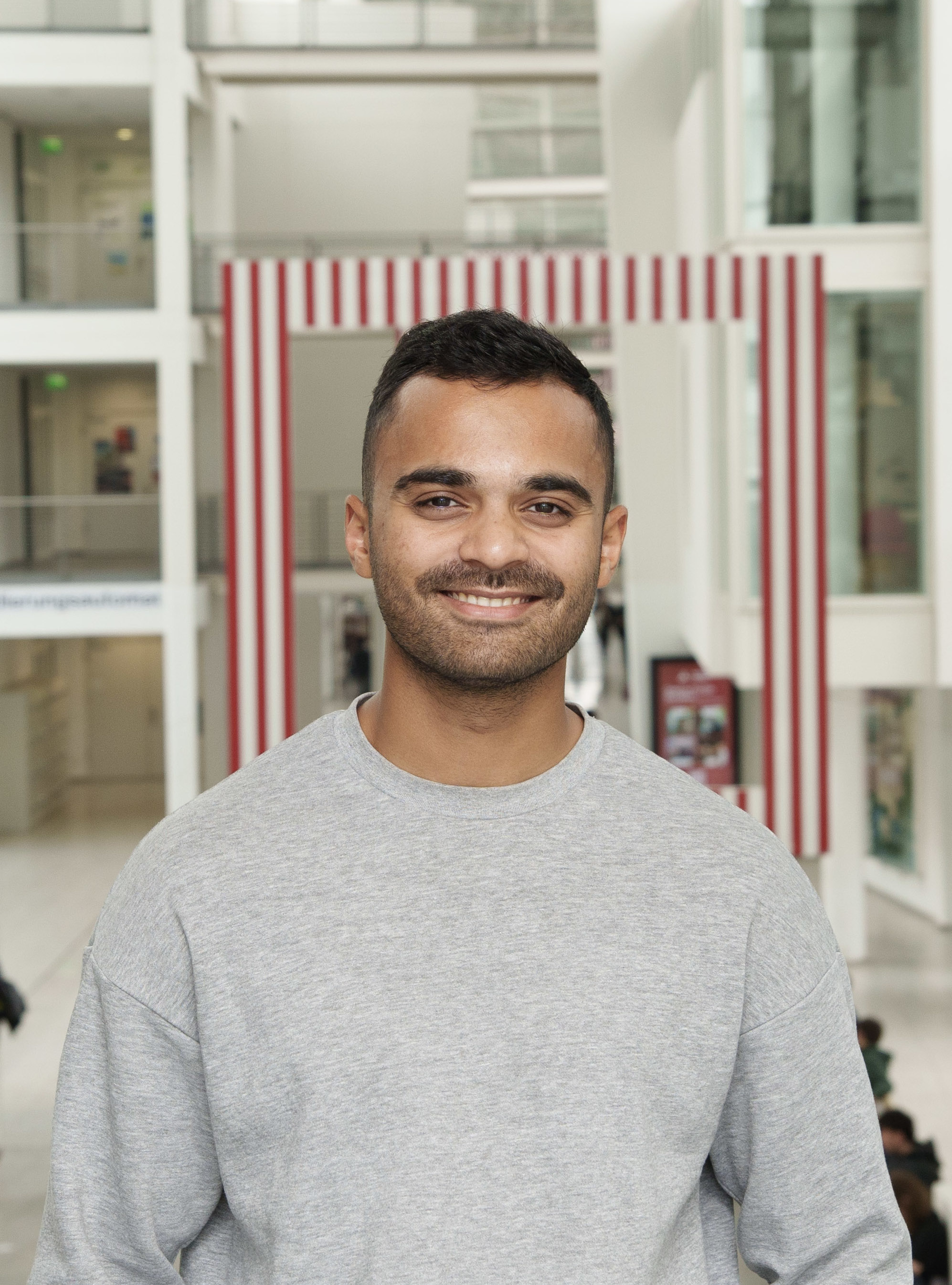}}]{Dheer Patel} obtained a B.Sc. degree in mechanical engineering and an M.Sc. degree in Robotics, Cognition, Intelligence from the Technical University of Munich in 2023 and 2025, respectively. He is currently pursuing a Ph.D. degree in automotive engineering with the Institute of Automotive Technology, Technical University of Munich with focus on controller evaluation for autonomous vehicles in urban and high-speed scenarios.
\end{IEEEbiography}
\begin{IEEEbiography}
[{\includegraphics[width=1in,height=1.25in,clip,keepaspectratio]{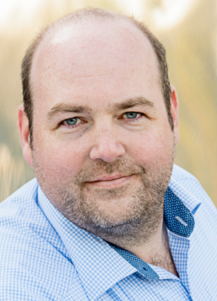}}]{Frank Diermeyer} received the Diploma and Ph.D. degrees in mechanical engineering from the Technical University of Munich, Germany, where he has been a Senior Engineer with the Chair of Automotive Technology, and the Leader of the Safe Operation Research Group, Autonomous Vehicle Lab since 2008. His research interests include teleoperated driving, human–machine interaction, and safety validation
\end{IEEEbiography}

\vspace{11pt}

\vfill

\end{document}